\documentclass[sigconf,screen,nonacm]{acmart}
\AtBeginDocument{%
  \providecommand\BibTeX{{%
    \normalfont B\kern-0.5em{\scshape i\kern-0.25em b}\kern-0.8em\TeX}}}

\usepackage{amsmath}
\usepackage{mathtools}
\usepackage{amsthm}
\usepackage{microtype}
\usepackage{graphicx}
\usepackage{subfigure}
\usepackage{subcaption}
\usepackage{booktabs} % for professional tables
\usepackage{tkz-tab}
\usepackage{colortbl}
\usepackage{wrapfig}
\usepackage{hyperref}
\usepackage{algorithm}
\usepackage{algorithmic}
\usepackage[capitalize,noabbrev]{cleveref}
\theoremstyle{plain}

\newtheorem{theorem}{Theorem}[section]
\theoremstyle{remark}
\newtheorem{remark}[theorem]{Remark}
\newcommand{\argmax}{\mathop{\mathrm{argmax}}}
\newcommand{\sbt}{\,\begin{picture}(-1,1)(-1,-3)\circle*{3}\end{picture}\ \,\,}
\begin{document}

%%
%% The ''title'' command has an optional parameter,
%% allowing the author to define a ''short title'' to be used in page headers.
\title[A Bandit Approach With Evolutionary Operators for Model Selection]{A Bandit Approach With Evolutionary Operators for Model Selection:  Application to Neural Architecture Optimization for Image Classification}

%%
%% The ''author'' command and its associated commands are used to define
%% the authors and their affiliations.
%% Of note is the shared affiliation of the first two authors, and the
%% ''authornote'' and ''authornotemark'' commands
%% used to denote shared contribution to the research.

\author{Margaux Brégère}
\authornote{Both authors contributed equally to this research.}
\email{margaux.bregere@edf.fr}
\affiliation{%
  \institution{
  EDF Lab Saclay, Palaiseau, France \\
  Univ. de Paris and Sorbonne Université, CNRS$\backslash$LPSM}
  \country{Paris, France}
}

\author{Julie Keisler}
\authornotemark[1]
\email{julie.keisler@edf.fr}
\affiliation{%
  \institution{EDF Lab Saclay, Palaiseau, France \\
  CNRS$\backslash$CRIStAL, Inria Lille, Université de Lille}
    \country{Villeneuve d’Ascq, France}
    }

%%
%% By default, the full list of authors will be used in the page
%% headers. Often, this list is too long, and will overlap
%% other information printed in the page headers. This command allows
%% the author to define a more concise list
%% of authors' names for this purpose.
\renewcommand{\shortauthors}{Brégère and Keisler}

%%
%% The abstract is a short summary of the work to be presented in the
%% article.
\begin{abstract}
This work formulates model selection as an infinite-armed bandit problem, namely, a problem in which a decision maker iteratively selects one of an infinite number of fixed choices (i.e., arms) when the properties of each choice are only partially known at the time of allocation and may become better understood over time, via the attainment of rewards.
Here, the arms are machine learning models to train and selecting an arm corresponds to a partial training of the model (resource allocation).
The reward is the accuracy of the selected model after its partial training.
We aim to identify the best model at the end of a finite number of resource allocations and thus consider the best arm identification setup. 
We propose the algorithm Mutant-UCB that incorporates operators from evolutionary algorithms into the UCB-E (Upper Confidence Bound Exploration) bandit algorithm introduced by~\citet{audibert2010best}.
Tests carried out on three open source image classification data sets attest to the relevance of this novel combining approach, which outperforms the state-of-the-art for a fixed budget.
\end{abstract}

%%
%% The code below is generated by the tool at http://dl.acm.org/ccs.cfm.
%% Please copy and paste the code instead of the example below.
%%
% \begin{CCSXML}
% <ccs2012>
%  <concept>
%   <concept_id>00000000.0000000.0000000</concept_id>
%   <concept_desc>Do Not Use This Code, Generate the Correct Terms for Your Paper</concept_desc>
%   <concept_significance>500</concept_significance>
%  </concept>
%  <concept>
%   <concept_id>00000000.00000000.00000000</concept_id>
%   <concept_desc>Do Not Use This Code, Generate the Correct Terms for Your Paper</concept_desc>
%   <concept_significance>300</concept_significance>
%  </concept>
%  <concept>
%   <concept_id>00000000.00000000.00000000</concept_id>
%   <concept_desc>Do Not Use This Code, Generate the Correct Terms for Your Paper</concept_desc>
%   <concept_significance>100</concept_significance>
%  </concept>
%  <concept>
%   <concept_id>00000000.00000000.00000000</concept_id>
%   <concept_desc>Do Not Use This Code, Generate the Correct Terms for Your Paper</concept_desc>
%   <concept_significance>100</concept_significance>
%  </concept>
% </ccs2012>
% \end{CCSXML}

% \ccsdesc[500]{Do Not Use This Code~Generate the Correct Terms for Your Paper}
% \ccsdesc[300]{Do Not Use This Code~Generate the Correct Terms for Your Paper}
% \ccsdesc{Do Not Use This Code~Generate the Correct Terms for Your Paper}
% \ccsdesc[100]{Do Not Use This Code~Generate the Correct Terms for Your Paper}

%%
%% Keywords. The author(s) should pick words that accurately describe
%% the work being presented. Separate the keywords with commas.

%% A ''teaser'' image appears between the author and affiliation
%% information and the body of the document, and typically spans the
%% page.
%\begin{teaserfigure}
%  \includegraphics[width=\textwidth]{sampleteaser}
%  \caption{Seattle Mariners at Spring Training, 2010.}
%  \Description{Enjoying the baseball game from the third-base
%  seats. Ichiro Suzuki preparing to bat.}
%  \label{fig:teaser}
%\end{teaserfigure}

%\received{20 February 2007}
%\received[revised]{12 March 2009}
%\received[accepted]{5 June 2009}

%%
%% This command processes the author and affiliation and title
%% information and builds the first part of the formatted document.
\maketitle
\section{Introduction}
\label{sec:intro}

% Problème de sélection de modèle générique

Accuracy of machine learning models significantly depends on some parameters which cannot be modified during training.
As the number of parameter combinations to be tested exponentially increases with the number of these parameters, it becomes costly and time-consuming to optimize them. 
Automating the selection of promising models, usually referred to as AutoML (Automated Machine Learning), is a fast-growing area of research (see~\citet{hutter2019automated} for a quite recent book).
We approach the model selection problem in a general manner without any restrictions on the nature of the hyper-parameters, such as the types of machine learning models, neural network architectures, or hyper-parameters of random forests. 
Our aim is to find the best model without making any assumptions about the task, model type, or reward to maximize. 
We assume that we have access to an infinite number of possible models and set a predetermined budget of resources $T$, used to train the models.
These resources are allocated to the models in the form of ``sub-trains'', such as iterations, data samples or features. 
The final (and hopefully the best) model is chosen by finding a good trade-off between exploration (training a large number of models) and exploitation (allocating a large budget to promising models). This process may fall under the umbrella of multi-armed bandits (see \citet{lattimore2020bandit} for an in-depth review).

% Présentation de MutantUCB
In this paper, we treat model selection as an instance of best-arm identification in infinite-armed bandits.
We propose a new model selection algorithm, called Mutant-UCB: an Upper Confidence Bound (UCB)-based algorithm (see~\citet{auer2002finite} for UCB's original idea) that incorporates a mutation operator from the evolutionary 
algorithms.  
This operator creates a new model from the neighborhood of the model selected by the bandit algorithm. Unlike most model selection algorithms, Mutant-UCB makes no assumptions about the solutions encoding, also called search space, or the reward function to be maximized, making it suitable for a wide range of configurations. 
The use of a UCB-type algorithm and adaptive resource allocation allows exploration of the search space, while the mutation operator effectively directs the search towards promising solutions. 
Results on a neural networks optimization problem demonstrate the relevance of this approach. 
%TODO : available framework #DRAGON

We begin this paper by presenting the setup of our bandit model selection approach in Section~\ref{sec:setup} and we position ourselves in relation to the state of the art.
Mutant-UCB, the algorithm we develop, is presented in Section~\ref{sec:MutantUCB}.
Section~\ref{sec:expe} is dedicated to experiments on the optimization of deep neural networks: we validate the performance of Mutant-UCB on three open-source image classification datasets.
Finally, Section~\ref{sec:conclu} discusses the advantages of Mutant-UCB compared to the state of the art and opens new research perspectives.

\section{Model selection problem setup: a bandit approach}
\label{sec:setup}

\subsection{Literature Discussion}
\label{subsec:stateofart}
Naive strategies for model selection are Grid or Random Search.
More sophisticated strategies address model selection as a sequential learning problem.
Two approaches stand out: 
\textbf{configuration selection} methods sequentially select new models (``close'' to promising models) to train, while \textbf{configuration evaluation} methods allocate more resources (training time) to promising models.
The first approach suggests that accuracy is regular with respect to some distance between models implying the existence of an underlying space. Therefore, two models that are close to each other will have similar performance. The second approach, on the other hand, makes no assumptions about the potential (smooth) links between model performances.

\paragraph{Evolutionary methods.} Among the configuration selection methods, evolutionary algorithms have been popular for many years (see, e.g., \citet{young2015optimizing} and \citet{ jian2023eenas}). Starting from an initial set of configurations, they evolve them towards performing models using unitary operators like the mutation (little change in the configuration). Even complex operators involving more than two configurations like the crossover are considered by~\citet{strumberger2019convolutional}.
These algorithms are highly versatile and can be applied to a wide range of setups.
The literature presents different methods that vary in terms of search spaces, i.e. the way configurations are encoded. The operators used to generate the new population typically depend on this encoding. 
Usually, the configurations are represented as character strings or lists and can be modified using bit-string mutations and combined with k-point crossovers (see \citet{eiben2015introduction} for more details). 
But recent works, mostly for neural networks architecture optimization, tried to design other representations and operators. For instance a tree-based mutation operator to optimize recurrent neural networks is proposed by \citet{rawal2018nodes}. \citet{awad2021dehb}  use differential evolutionary operators to optimize neural network hyper-parameters and architectures.
One disadvantage of the evolutionary algorithms is the large number of parameters involved, such as the population size, the selection function, or the elitism rate. Choosing the appropriate values for these parameters can be complex.

\paragraph{Bayesian optimization.} Bayesian optimization has recently emerged as a more efficient approach than evolutionary methods in AutoML (see, among others, \citet{malkomes2016bayesian}, \citet{ zoph2016neural} and \citet{kandasamy2018neural}). It is a sequential optimization technique commonly used to minimize black-box functions. Those algorithms are based on two main components, a surrogate model that approximates the unknown black-box function, and an acquisition function that selects the next element in the search space to be evaluated. 
One major limitation of these acquisition functions is their reliance on strong assumptions about the  black-box function and the search space (see \citet{garrido2020dealing} for further details). Therefore, we did not employ a Bayesian optimization algorithm in our experiments as we aimed to avoid making any assumptions about the smoothness, distance or continuity of the search space or the reward function.

\paragraph{Bandits approaches.} Firstly, still in the field of Bayesian optimization, the extensions GP-UCB and KernelUCB (see, \citet{srinivas2009gaussian} and \citet{valko2013finite}, respectively) of the classical UCB bandit algorithm and more recent algorithms largely inspired by them (see, e.g., \citet{dai2023quantum}) have been massively used for optimization and eventually model selection.
The BayesGap algorithm introduced by \citet{hoffman2014correlation} connects Bayesian optimization approaches and best arm identification, assuming correlations among the arms.
More recently, \citet{huang2021neural} sees the neural architecture search as a combinatorial multi-armed bandit problem which allows the decomposition of a large search space into smaller blocks where tree-search methods can be applied more effectively and efficiently.
Configuration evaluation approaches have also been investigated in an infinite or multi-armed bandit framework.
At each iteration of the algorithm, a new arm/model can be drawn from an infinite search space containing the models and added to the set of models already (more or less) trained. 
\citet{karnin2013almost} proposes the Sequential (or Successive) Halving algorithm, which splits the given budget evenly across an optimal number of elimination rounds, and within a round, pulls arms in a uniform manner.
It comes with solid theoretical guarantees that have recently been improved by \citet{zhao2023revisiting}.
\citet{li2018Hyperband} proposes the Hyperband algorithm, a robust extension of Sequential Halving, and applies it to deep neural networks hyperparameters optimization.
Moreover, \citet{shang2019simple} introduces D-TTTS, an algorithm inspired by Thompson sampling.
Hybrid methods combine adaptive configuration selection and evaluation: \citet{terayama2021black} proposes a rule to stop training a model prematurely based on the predicted performance from Gaussian Process;
in addition, \citet{kandasamy2016gaussian} extends GP-UCB to enable sequential model training (and thus resource allocation).

\paragraph{Best-arm identification in infinite-armed bandits.}
The stochastic infinite-armed bandit framework has been introduced and studied for the cumulative reward maximization problem by \citet{berry1997bandit} and \citet{wang2008algorithms}.
\citet{carpentier2015simple} and \citet{aziz2018pure} study best armed identification problem in this framework.
Theoretical results attest to the performance of their strategies (SiRI and extensions; $\alpha$,$\epsilon$-KL-LUCB, respectively).

\subsection{Contributions}
The main contribution of this study is the Mutant-UCB algorithm, which incorporates operators from evolutionary algorithms into the UCB-E (Upper Confidence Bound Exploration) algorithm introduced by~\citet{audibert2010best}.
It combines both configuration evaluation and configuration selection approaches: 
it is sequential in computation and picks a (generally promising) model thanks to a UCB-based criteria. 
Then it either continues its training (resource allocation) or creates and starts training a new model derived from the selected one thanks to the ``mutation'' operation of an evolutionary algorithm. 
This last possibility is based on the intuition that the expected ``mutant model'' accuracy will be close to that of the original model.
While bandit approaches have been used to design the ``selection'' operator for evolutionary algorithms by~\citet{li2013adaptive}, to our knowledge this is the first time that operators from evolutionary algorithms are incorporated into a bandit algorithm. 

Afterwards, we compare Mutant-UCB to a Random Search, the evolutionary algorithm proposed by~\citet{keisler2023algorithmic} and the Hyperband algorithm introduced by~\citet{li2018Hyperband} on three open source data sets collected for image classification: CIFAR-10 \citep{krizhevsky2009learning}, MRBI \citep{larochelle2007empirical} and SVHN \citep{netzer2011reading}. 
For a fair comparison, Mutant-UCB and the evolutionary algorithm under consideration share the same mutation operation. 

\subsection{Set-up}
In the infinite-armed bandit framework, when a new arm $k$ is pulled from the reservoir, the expectation of the accuracy of the associated model $\mu_k$ pulled from the search space is assumed to be an independent sample from a fixed distribution.
With $T$ a fixed budget, at each round $t= 1,\dots,T$, an arm $I_t$ is picked and a sub-train (allocation of a resource) is performed on the associated model.
This model is then evaluated on a validation data set $\mathcal{D}_{\mathrm{valid}}$ using an accuracy function $\mathrm{acc}$ we aim to maximize. 
This accuracy corresponds to the reward $a_t$.  
After $T$ rounds, we select the final arm $\widehat{I}_T$.
In what follows, the untrained model associated with arm $k$ is denoted $f_k$, and after $N_k$ sub-trains we denote it $f_k^{N_k}$.

\section{A UCB-based algorithm incorporating mutation operators from evolutionary algorithms}
\label{sec:MutantUCB}

\subsection{A brief reminder of the UCB-E algorithm}
\label{subsec:UCBE}
Designed for best arm-identification in a $K$-multi-armed bandit problem, the UCB-E algorithm proposed by\citet{audibert2010best} is recalled in \cref{alg:EUCB}. 
This highly exploratory policy is based on the principle of optimism in the face of uncertainty, in the spirit of the UCB algorithm introduced by \citet{auer2002finite}.
It aims to find the best model among $K$ untrained models $f_1,\dots,f_K$ sampled from the search space.
The algorithm starts by $K$ rounds of deterministic exploration: it performs a first sub-train per model and observes the accuracies $a_k = \mathrm{acc}\big(f_{k}^{1}, \,\mathcal{D}_{\mathrm{valid}}\big)$. 
At each round $t=K+1,\dots, T$, and for each $k$, it computes the empirical mean accuracy $\hat{\mu}_{k,t}$ from the previous rewards associated with arm $k$:
\begin{equation}
    \label{eq:mu:N}
    \hat{\mu}_{k,t} = \frac{1}{N_{k,t}} \sum_{s=1}^{t-1} a_t \mathbf{1}_{I_s = k} \, \quad \mathrm{with} \quad {N_{k,t}} = \sum_{s=1}^{t-1} \mathbf{1}_{I_s = k }\,.
\end{equation}
Then, it chooses the arm optimistically:
\begin{equation}
    \label{eq:algo}
    I_t \in \argmax_{k \in \{ 1,\dots K\}}{ \bigg\{ \, \hat{\mu}_{k,t} + \sqrt{ \frac{E}{N_{k,t}}} \,\bigg\}} \,,
\end{equation}
performs a sub-train on the associated model and receives the reward $a_t = \mathrm{acc}\big(f_{I_t}^{N_{I_t,t+1}}, \,\mathcal{D}_{\mathrm{valid}}\big)$.
For the sake of readability, the iteration index for the counting $N_k$ and empirical mean $\mu_k$ variables in Algorithms~\ref{alg:EUCB} and~\ref{alg:MutantUCB} have been removed. These variables are updated throughout the iterations.

The core issue is the tuning of the exploration parameter $E$. \citet{audibert2010best} show that the optimal value depends on the difficulty of the underlying bandit problem, which has no reason to be known in advance.

\begin{algorithm}[tb]
   \caption{$UCB-E$}
   \label{alg:EUCB}
\begin{algorithmic}
\STATE \textbf{Inputs}: 
\STATE \quad $T$ budget
\STATE \quad $E$ exploration parameter 
\STATE \quad $K$ number of untrained models 
\STATE \textbf{Initialization}  
\STATE \quad Sample $K$ untrained models $f_1,\dots, f_K$
\STATE \quad \textbf{For} $k =1,2,\ldots, K$ 
\STATE \quad \quad Perform a first sub-train on $f_k$ which becomes $f_k^1$
\STATE \quad \quad Get the reward $a_k = \mathrm{acc}\big(f_k^1, \mathcal{D}_{\mathrm{valid}}\big)$ 
\STATE \quad \quad Define $N_{k} =  1$, $\hat{\mu}_k = a_k$
\STATE \textbf{For} $t=K+1,K+2,\ldots, T$ 
\STATE \quad Choose $I_t \in \argmax_{k \in \{ 1,\dots K\}}{ \Big\{\hat{\mu}_k + \sqrt{ \frac{E}{N_k}}\Big\}}$
\STATE \quad  Perform a sub-train: model $f_{I_t}^{N_{I_t}} $ becomes $f_{I_t}^{N_{I_t} + 1}$ 
\STATE \quad   Get the reward $a_t = \mathrm{acc}\big(f_{I_t}^{N_{I_t} + 1}, \,\mathcal{D}_{\mathrm{valid}}\big)$ 
\STATE \quad  Update $\hat{\mu}_{I_t} =\frac{1}{N_{I_t}+1} \big( a_t + N_{I_t}\hat{\mu}_{I_t}\big)$ and  $N_{I_t} =N_{I_t} +1$ 
\STATE \textbf{Output}: 
\STATE  \quad Model $f_{\widehat{I}_T}^{N_{\widehat{I}_T}}$  where
$\widehat{I}_T \in \argmax_{k \in \{ 1,\dots K\}}{\hat{\mu}_k}$  
\end{algorithmic}
\end{algorithm}

\subsection{Main contribution: the Mutant-UCB algorithm}
\label{subsec:MutantUCB}
Mutant-UCB, presented in \cref{alg:MutantUCB}, incorporates two main ideas into UCB-E.
First of all, there is no point in multiplying the number of sub-trains for the same model: there generally comes a time when it is no longer useful, so we can potentially define a maximum number of sub-trains $N$.
Note that this idea of a maximum quantity of resources that can be allocated to a single model was already present in the Hyperband algorithm proposed by~\citet{li2018Hyperband}.
Furthermore, in model selection, it is not uncommon for similar models to perform similarly, which is why configuration selection methods may be so effective in this task. 
In general, the main problem lies in defining a distance between models: search spaces are usually high-dimensional and hyper-parameters are of various kinds (learning rate, type of activation function, number of neurons, etc.).
While the notion of distance between two models is not easy to define, evolutionary algorithms offer a good compromise: they breed new individuals through crossover and mutation operations.
Crossover operations mix two individuals, while mutation operations can be applied to a single individual in order to create ``mutants'', involving some tiny changes. Those ``mutants'' can be seen as neighbors of the initial point. 
We could therefore imagine that a model chosen by the algorithm could mutate to give rise to a new one, with the intuition that the mutant and its original model will have similar accuracies.
To our knowledge, the inclusion of mutation operators of evolutionary algorithms in a bandit algorithm is completely new.

\begin{algorithm}[tb]
   \caption{Mutant-UCB}
   \label{alg:MutantUCB}
\begin{algorithmic}
\STATE \textbf{Inputs}: 
\STATE \quad $T$ budget
\STATE \quad $E$ exploration parameter 
\STATE \quad $K$ initial number of models 
\STATE \quad $N$ maximum number of sub-trains that can be allocated 
\STATE \quad $\quad$ to a single model
\STATE \textbf{Initialization}  
\STATE \quad Sample $K$ untrained models $f_1,\dots, f_K$
\STATE \quad \textbf{For} $k =1,2,\ldots, K$ 
\STATE \quad \quad Perform a first sub-train on $f_k$ which becomes $f_k^1$
\STATE \quad \quad Get the reward $a_k = \mathrm{acc}\big(f_k^1, \mathcal{D}_{\mathrm{valid}}\big)$ 
\STATE \quad \quad Define $N_k = \overline{N}_k = 1$, $\hat{\mu}_k = a_k$
\STATE \textbf{For} $t=K+1,K+2,\ldots, (T-N+1)$ 
\STATE \quad Choose $I_t \in \argmax_{k \in \{ 1,\dots K\}}{ \Big\{\hat{\mu}_k + \sqrt{ \frac{E}{N_k}}\Big\}}$
\STATE \quad Sample $X_t \sim \mathcal{B}(p_t)$ with $p_t = 1- \overline{N}_{I_t}/N$
\STATE \quad  \textbf{If} $X_t = 1$:
\STATE \quad \quad  Perform a sub-train: 
model $f_{I_t}^{\overline{N}_{I_t}} $ becomes $f_{I_t}^{\overline{N}_{I_t} + 1}$ 
\STATE \quad \quad  Get the reward $a_t = \mathrm{acc}\big(f_{I_t}^{\overline{N}_{I_t} + 1}, \,\mathcal{D}_{\mathrm{valid}}\big)$ 
\STATE \quad \quad  Update $\hat{\mu}_{I_t} =\frac{1}{\overline{N}_{I_t}+1} \big( a_t + \overline{N}_{I_t}\hat{\mu}_{I_t}\big)$,  $N_{I_t} =N_{I_t} +1$ 
\STATE \quad \quad  and $ \overline{N}_{I_t} = \overline{N}_{I_t} + 1$
\STATE \quad  \textbf{Else} :
\STATE \quad \quad  Update the number of models $K = K +1$
\STATE \quad \quad  Create a mutant model $f_{K}$ from $f_{I_t}^{\overline{N}_{I_t}}$
\STATE \quad \quad  Perform a first sub-train on $f_{K}$ which becomes $f_K^1$
\STATE \quad \quad Get the reward $a_t = \mathrm{acc}\big(f_K^1, \mathcal{D}_{\mathrm{valid}}\big)$ 
\STATE \quad \quad  Define $N_K = \overline{N}_K = 1$, $\hat{\mu}_K = a_t$
\STATE \quad \quad  Update $N_{I_t} = N_{I_t} +1$ 
\STATE \textbf{Finalization}  
\STATE \quad Select the best model $\widehat{I}_T \in \argmax_{k \in \{ 1,\dots K\}}{\hat{\mu}_k}$
\STATE \quad Finalize its training by performing $N-\overline{N}_{\widehat{I}_T}$ sub-trains
\STATE \textbf{Output}: $f_{\widehat{I}_T}^N$  
\end{algorithmic}
\end{algorithm}

Like the UCB-E algorithm, Mutant-UCB starts with the first sub-train of $K$ models.
At each round $t = K+1, \dots $, it still chooses the next arm optimistically, by resolving Equation~\eqref{eq:algo}.
For an arm $k$, we recall that $N_{k,t}$ is the number of times the arm has been picked before round $t$ - see Equation~\eqref{eq:mu:N}. 
We now introduce $\overline{N}_{k,t}$, the integer that counts the number of times the model associated with arm $k$ has been trained.
Once arm $I_t $ is picked, with $p_t = 1- \overline{N}_{I_t,t}/N$:

\sbt a sub-train is performed on $f_{I_t}^{\overline{N}_{I_t,t}}$ with probability $p_t$ or

\sbt a mutation is performed on $f_{I_t,t}^{\overline{N}_{I_t,t}}$ with probability $1-p_t$.

The mutation is performed on the trained model $f_{I_t}^{\overline{N}_{I_t,t}}$ -~and not just $f_{I_t}$~- to include the case where certain parameters of the model optimized during training (e.g., weights of neuron networks) are passed on to its mutant.
We detail the mutation operation for our use-case in Section~\ref{sec:expe}. 
When a mutation occurs, a new model is created, a first sub-train is performed and the model is added to the list of potential models to be retained at the end of the algorithm.
Thus, the number of models $K$ increases by one each time a mutant model is created. 

\begin{remark}
\label{req:ucbmutant}
  When a new model comes into play, it is very likely to be quickly chosen by the algorithm, even if its accuracy is not good: the algorithm must explore this new possibility. 
  The ``sleeping bandit'' framework, in which new arms may be added and/or become unavailable during the algorithm execution, is studied by \citet{kleinberg2010regret}.
  It proposes a very natural extension of UCB: the Awake Upper Estimated Reward algorithm and shows there is no need to adapt the confidence bounds. 
\end{remark}

The probability $p_t$ decreases as the model goes along its sub-trains and guarantees that it will not be trained more than $N$ times. 
The more the model has been trained, the more likely it is to mutate when selected. 
The underlying idea is that further training will probably have little effect or even over-fit in the case of neural networks, and that if the algorithm selects this already well-trained model, it is because it may have good accuracy (and it will probably be the same for a mutant model).
Note that the probability $p_t$ is linear in $N_{I_t,t}$; this choice is arbitrary and we could quite easily have chosen another type of relationship, e.g., $p_t = 1- \exp(N_{I_t,t}-N)$.
The algorithm ends with a finalization phase: the best model, in terms of average accuracy, $\widehat{I}_T$ is selected among the initial models and the mutant models and its training is completed with $N-\overline{N}_{\widehat{I}_{T},T-N+2}$ additional sub-trains.

\section{Experiments}
\label{sec:expe}

In this section, we evaluate the performance of the Mutant-UCB algorithm, on neural networks optimization. 
In order to highlight the advantages of our method, we put ourselves in a case where we make no assumptions about the smoothness of the reward function $\mathrm{acc}$ and we do not consider any distance between the elements $f_k$ from our search space. We therefore compare our methods with three algorithms that are applicable in this case: a Random Search, the Hyperband algorithm and an evolutionary algorithm. This neural networks optimization is applied to three image classification data sets.

\subsection{Experiment design}
\label{subsec:expedesign}

\paragraph{Data sets.}
We performed our experiments using three image classification data sets, also used by \citet{li2018Hyperband} to introduce the Hyperband algorithm: CIFAR-10~\cite{krizhevsky2009learning}), Street View House Numbers (SVHN~\cite{netzer2011reading}) and rotated MNIST with background images, also called MRBI~\cite{larochelle2007empirical}.
These first two data sets contain $32 \times 32$ RGB images, while MRBI contains $28 \times 28$ gray-scale images. The labels for each data set are converted to integers between 0 and 9. We split each data set into a training, a validation and a testing set. 
The training set is used to optimize the model weights  (namely to perform the sub-trains), while the validation set is used to evaluate the configurations in the selection model algorithms (i.e. to get the rewards). Finally, the accuracies of the configuration selected by the algorithms are computed on the testing set to assess their quality.
CIFAR-10 has 35k image on the train set, 15k on the validation set and 10k on the test set, SVHN has 51k, 22k and 26k and MRBI 10k, 2k and 50k data points on the three sets respectively. For all data sets we standardized the images so the input has a mean of zero and a standard deviation of one.

\paragraph{Search space: the pool of possible configurations.}
We used the DRAGON framework developed by
\citet{keisler2023algorithmic} to encode our neural networks. Their article contains an explanation and a tutorial on how to use the associated implemented package.
In this framework, neural networks are represented as directed acyclic graphs (DAGs), where the nodes represent the layers (e.g., recurrent, feed-forward, convolutional) and the edges represent the connections between them.
The task on which we want to try our algorithms is image classification.
To do so, we define a generic search space (the pool of possible configurations $f_k$) with DRAGON, dedicated to the task at hand.
Any sampled configuration $f_k$ will be made of two directed acyclic graphs. The first one processes 2D data, and can be made of 2D convolutions, 2D pooling, normalization and dropout layers. 
The second one consists in a flatten layer followed by MLPs (Multi-Layers Perceptrons) and normalization layers.
A final MLP layer is added at the output of the model to convert the latent vector into the desired output format. 
The framework includes operators, namely mutations and crossovers to modify and thus optimize the graphs.  
The mutation operators modify the neural network architecture by adding, removing or modifying the nodes and the connections in the graph. 
They can also be applied within the nodes, on the neural network hyper-parameters (e.g., convolution layer kernel size or an activation function). Crossover involves exchanging parts of two graphs.

\paragraph{Sub-trains.}We trained our neural networks using a cyclical learning rate, as proposed by \citet{huang2017snapshot}. When the learning rate is low, the neural network reaches a local minimum. Right after, the learning rate goes up again taking the model out of the local minimum. We consider in our experiments that a sub-train is one of this loop, with learning rate getting from its maximum to its minimum. We let $N$ be the maximum number of sub-trains for a given element $f_k$ from our search space.

\paragraph{Baselines.}
Random Search, the Evolutionary Algorithm (EA), Hyperband and Mutant-UCB have all been implemented so that they can be used with the DRAGON framework. 
They all use the same training and validation functions to assess the neural networks performance, and share a common budget, namely $T$. 
For Random Search, we randomly select $K_{\mathrm{\textsc{RS}}} = T/N$ neural networks.
For each of them we perform $N$ sub-trains, resulting in $T$ sub-trains in total. 
For the Evolutionary Algorithm, we implemented an asynchronous (or steady-state) version. 
Compared to the standard algorithm, the steady-state evolutionary algorithm of ~\citet{liu2018hierarchical} enhances efficiency on High-Performance Computing (HPC) by producing two offsprings from the population as soon as a free process is available, rather than waiting for the entire population to be evaluated. We set an initial population of size $K_{\mathrm{\textsc{EA}}}$, where the deep neural networks are randomly initialized. 
We perform $N$ sub-trains on each of these models. 
Then, we evolve the population using the mutation and crossover operators from the DRAGON framework. 
If a generated offspring is better than the worst model from the population, it replaces it. 
During the optimization procedure, we generate $T/N - K_{\mathrm{\textsc{EA}}}$ offsprings and we perform $N$ sub-trains on each, resulting in a total of $T$ sub-trains. 
For Hyperband we ran the algorithm with its parameters $R$ and $\eta$ such that the total number of sub-trains is $T$ and that each model can be trained only $N$ times (see, \citet{li2018Hyperband} for further details).
%The algorithms UCB-E and Mutant-UCB respectively start with populations of size $K_{\mathrm{\textsc{UCB-E}}}$ and $K_{\mathrm{\textsc{Mutant}}}$, and run with a budget of $T$. For a fair comparison, EA and Mutant-UCB use the same mutation operators.
The algorithm Mutant-UCB starts with an initial population of $K_{\mathrm{\textsc{Mutant}}}$ and runs with a budget of $T$. For a fair comparison, EA and Mutant-UCB use the same mutation operators.
We set $K_{\mathrm{\textsc{EA}}}  << K_{\mathrm{\textsc{Mutant}}} \lesssim K_{\mathrm{\textsc{RS}}}$. 
%<< K_{\mathrm{\textsc{UCB-E}}}
Indeed, as each configuration is fully trained in the evolutionary algorithm, $K_{\mathrm{\textsc{EA}}}$ must be much lower than $T/N$ to allow the creation of a sufficient number of offsprings.
%For UCB-E, taking $ T/N =  K_{\mathrm{\textsc{RS}}}  << K_{\mathrm{\textsc{UCB-E}}}$ permits a large exploration.
%With the mutation operator we create new configurations during the optimization procedure, so there is no need in taking as many individuals in the initial pool for Mutant-UCB than for UCB-E.
%This is why we set $K_{\mathrm{\textsc{Mutant}}}$ smaller than $K_{\mathrm{\textsc{UCB-E}}}$.
Similarly, Mutant-UCB mutation operator will create new configurations during the optimization procedure. However the evolutionary algorithm will generate even more individuals with the crossover, so $ K_{\mathrm{\textsc{Mutant}}}$ may be higher than $K_{\mathrm{\textsc{EA}}}$.
We then set $ K_{\mathrm{\textsc{Mutant}}}$ a bit smaller than $T/N$.
We emphasize that, with the creation of offsprings and mutants, the final number of evaluated models by the evolutionary algorithm and Mutant-UCB will be much higher than $K_{EA}$ and $K_{\mathrm{\textsc{Mutant}}}$, respectively.
In addition, Random Search and the evolutionary algorithm fully train each configuration tested (of which there are $T/N$), while Hyperband and Mutant-UCB allow some of them to be partially trained (resulting in a final population of more than $T/N$ configurations). 
%As the mutation operation is a form of exploration, we reduce the exploration parameter $E$ for Mutant-UCB compared to UCB-E.

\subsection{Results}
\label{subsec:res}

\begin{table}
\caption{Number of tested models and accuracies (in \%) of the best model for Random Search (RS), asynchronous evolutionary algorithm (EA) and Mutant-UCB on CIFAR-10, MRBI and SVHN data sets.}
\label{tab:results}
\vskip 0.15in
\begin{center}
\begin{footnotesize}
\begin{tabular}{lcccr}
\toprule
Data set & CIFAR-10 & MRBI & SVHN \\
\midrule
RS  &  1~000 $\cdot$ 75.3 & 1~000 $\cdot$ 75.5  & 1~000 $\cdot$ 90.7  \\
EA  &  1~000 $\cdot$ 77.1 &   1~000 $\cdot$ 79.5  & 1~000 $\cdot$ 91.9\\
Hyperband  & 2~400 $\cdot$ 75.4  & 2~400 $\cdot$ 75.9 & 2~400 $\cdot$ 91.0    \\
%UCB-E    & -  & -  &  - \\
Mutant-UCB   & \textbf{ 3~399 }$\cdot$\textbf{ 79.5\,} &  \textbf{ 3~463} $\cdot$ \textbf{80.5\,} &  \textbf{ 3~471} $\cdot$ \textbf{92.4\,}  \\
\bottomrule
\end{tabular}
\end{footnotesize}
\end{center}
\vskip -0.1in
\end{table}

\begin{figure}
\centering
\renewcommand\thefigure{(\alph{figure})}
\begin{subfigure}{}
    \includegraphics[width=0.45\textwidth]{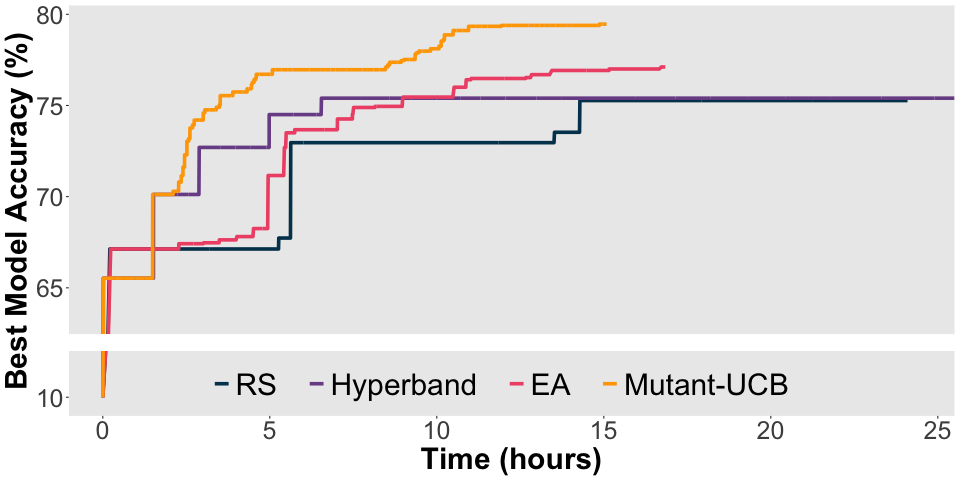}
    \vspace*{-4mm}
    \caption{CIFAR-10.}
    \label{fig:cifar}
\end{subfigure}
\hfill
\begin{subfigure}{}
    \includegraphics[width=0.45\textwidth]{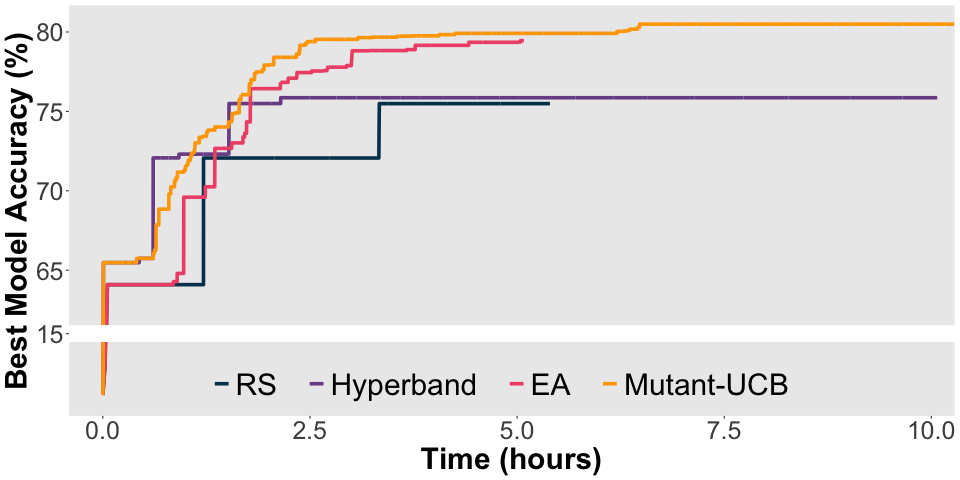}
    \vspace*{-4mm}
    \caption{MRBI.}
    \label{fig:mrbi}
\end{subfigure}
\hfill
\begin{subfigure}{}
    \includegraphics[width=0.45\textwidth]{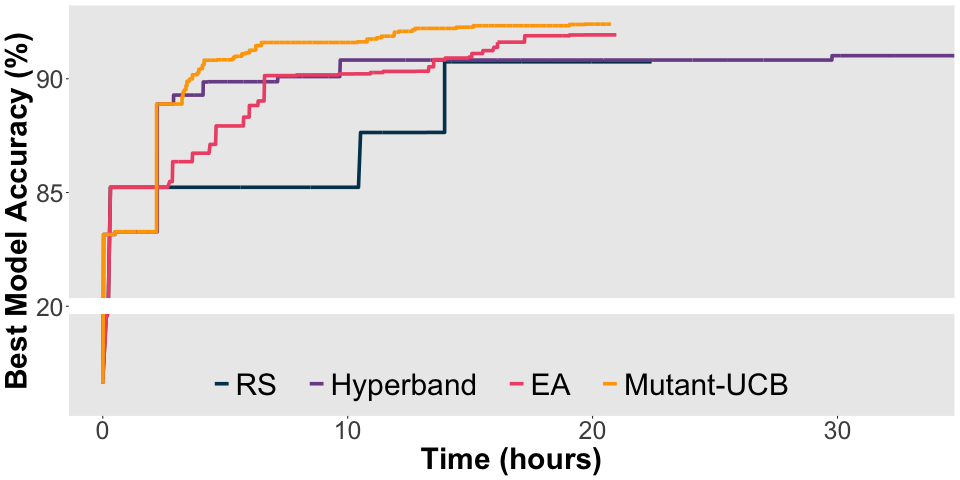}
    \vspace*{-4mm}
    \caption{SVHN.}
    \label{fig:SVHN}
\end{subfigure}
\setcounter{figure}{1}
\renewcommand\thefigure{\arabic{figure}}
\caption{Accuracy of the best model over computational time for Random Search (RS), asynchronous evolutionary algorithm (EA) and Mutant-UCB on CIFAR-10, MRBI and SVHN data sets.}
\label{fig:accuracies}
\end{figure}

We run the experiments with $T=10,000$, $N=10$ and $E=0.05$ for Mutant-UCB.
% TODO : experiments on MINST for mutant ?
The tuning of the parameter $E$ for Mutant-UCB is not as important as for the UCB-E algorithm. 
We only need the algorithm to not explore too much, since the mutation operator represents an additional form of exploration. 
Indeed, in the UCB-E algorithm, parameter $E$ is responsible for managing the balance between exploration (through the execution of a few sub-trains for numerous models) and exploitation (through the execution of numerous sub-trains for a few models).
Here, the Mutant-UCB algorithm incorporates two forms of exploration: firstly, through the sub-training of numerous models (which is also linked to the initial number of models $K$), and secondly, through the generation of mutants. 
If $E$ is relatively large, numerous models will be tested, with a correspondingly smaller number of mutants generated (such an algorithm would be similar to Random Search). Conversely, if $E$ is relatively small, a limited number of models will be fully trained, resulting in a substantial number of mutants being generated from them (such an algorithm would be similar to an evolutionary algorithm without the crossover operator, selecting only the best models at each iteration).

Each sub-trains contains 10 epochs, resulting in a maximum of 100 training epochs, and the learning rate is set to 0.01. Each experiment is run on a HPC environment using 20 NVIDIA V100 GPUs.

We display Table~\ref{tab:results} the maximum accuracies and the number of tested models for each algorithm from our baseline. We see that Mutant-UCB outperforms Random Search, the evolutionary algorithm and Hyperband for every data sets. The use of the mutation operator seems to be the primary factor in this performance. Indeed, the evolutionary algorithm comes well ahead of Hyperband and Random Search. 
It would seem that digging around promising solutions leads to better configurations. 
However, resources allocation also seems to be a key factor. 
Hyperband is in fact slightly better than Random Search, and converges much faster, as it can be seen in Figure~\ref{fig:accuracies}.
The computation times to perform the $T$ iterations vary a lot between the algorithms and the task at hand. 
The hardware definitely has an impact on this, but the resources allocation will also play a crucial role. 
Throughout the paper there is an implicit assumption that a sub-train's budget in terms of memory and time is independent of the model, which is not entirely correct.
Indeed, performing more sub-trains on configurations which are more complex take generally a longer time and affect the total duration of the experiment.
Mutant-UCB, with both the mutation operator and the resources allocation has the fastest convergence and yields the better accuracies. Appendix~\ref{app:architecture} details the models found by the different algorithms.

Codes are available in the supplementary material.

\section{Discussion}
\label{sec:conclu}

This work presents Mutant-UCB, an innovative model selection algorithm, which combines a UCB-based bandit algorithm with evolutionary algorithm operators. Most configuration selection approaches, such as Bayesian optimization or continuous bandit algorithms, typically consider a normed vector space to represent the pool of possible configurations. 
These approaches assume that the reward function is smooth, meaning that two configurations that are close in the underlying vector space lead to close accuracies.
Mutant-UCB and the other algorithms in the baseline do not require any smoothness assumptions.
Besides, thanks to its resource allocation, Mutant-UCB demonstrates a high exploratory potential. It can evaluate more models within a similar budget compared to Random Search or evolutionary algorithms. For example, on the MRBI data set, with a budget $T=10,000$, Mutant-UCB evaluated $3,500$ configurations, while the Evolutionary Algorithm and Random Search only evaluated $1,000$.
The use of a mutation operator, on the other hand, reinforces the exploitation of promising solutions and allows us to reach much higher performance configurations than Hyperband and Random Search.  
The mutation can be viewed as a concept of proximity: a mutant and its original model are close together, as defined by the chosen operator (which does not require any normed vector space).
It remains more permissive than the operators of the evolutionary algorithm. 
In particular, the crossover from the evolutionary algorithm requires homogeneity between the elements of the search space, unlike Mutant-UCB. 
Thus, Mutant-UCB could be used with a search space that combines various machine learning models, such as neural networks, random forests, or boosting; as soon as we define a mutation operator for each type of model.
Finally, the Mutant-UCB algorithm is highly scalable in an HPC environment because configurations are evaluated independently and asynchronously, in contrast to Hyperband and classical evolutionary algorithms, which evaluate populations synchronously. 
In summary, Mutant-UCB has several advantages that make it an attractive algorithm, in addition to its baseline-beating performance demonstrated in the previous section. 
One disadvantage is the need to store the weights of all previous configurations evaluated.
This is because the pool of solutions on which we apply the UCB part of the algorithm is not limited, unlike the other algorithms from the baseline.

\paragraph{Prospects.}
The experiments demonstrate the relevance of Mutant-UCB. A challenge for further work would be to obtain an upper bound on the simple regret of this algorithm to attest a theoretical performance.
In this best-arm identification problem for infinite-armed bandit framework; the simple regret will be the accuracy of the best possible model minus the accuracy of the model selected by the algorithm.
To obtain theoretical results, we believe that it will be necessary to introduce some concepts from sleeping bandits due to the creation of mutants (see, e.g.,~\citet{kleinberg2010regret} and contextual bandits to model the proximity between mutants and their original models (see, among other \citet{li2010contextual}). 
One of the most challenging aspects of the analysis will be to select an appropriate hypothesis regarding the distribution of rewards conditionally to the chosen arm. The classical stochastic bandit assumptions are not applicable in this context, as they suggest that,  conditionally to the chosen arm, the accuracy does not depend on the number of sub-trains performed. However, empirical evidence indicates that performing multiple sub-trains enhance performance.
Furthermore, in order to legitimize the idea of integrating mutation operations, and hopefully get the simple regret bound, it seems essential to add an assumption about the distribution of the accuracies of mutants.
It should also be noted that the HPC environment and the neural network training duration puts us in a context where rewards arrive with a delay (namely, in a delayed bandits framework - see, e.g., \citet{vernade2020linear}) and not all arms are available at all times (sleeping bandits again).

The strategy for creating mutant models is independent of the choice of bandit algorithm used to select the arms. Alternative approaches, other than those based on UCB, could be considered.

In this paper we applied Mutant-UCB to a very generic problem: neural networks optimization for image classification. The flexibility of this algorithm means that it can be applied to a wide range of problems. A natural extension of this paper would be to apply Mutant-UCB to a variety of tasks, models and search spaces where state-of-the-art algorithms, by their very nature, would be limited or even unusable.

\bibliographystyle{ACM-Reference-Format}
\bibliography{sample-base}

\appendix
\section{Models found by the algorithms}
\label{app:architecture}

We set the same general seed for each algorithm and data set: CIFAR-10, MRBI and SHVN. This means that the initial pool of solutions is exactly the same for each algorithm and for the three data sets. Hyperband and Random Search found exactly the same architecture, indicating that the best configuration was at the beginning of this pool. Indeed, despite the fact that Hyperband looks at more than twice as many configurations as Random Search, it resulted in the same configuration as Random Search. Hyperband's best results come from the fact that 10 sub-trains are probably too many for this model, which overfits in the case of Random Search. More surprisingly, the best model for Hyperband and Random Search, shown in Figure~\ref{fig:graph_rs_hb}, is identical for all three data sets.

\begin{figure}[h!]
    \centering
    \includegraphics[width=5cm]{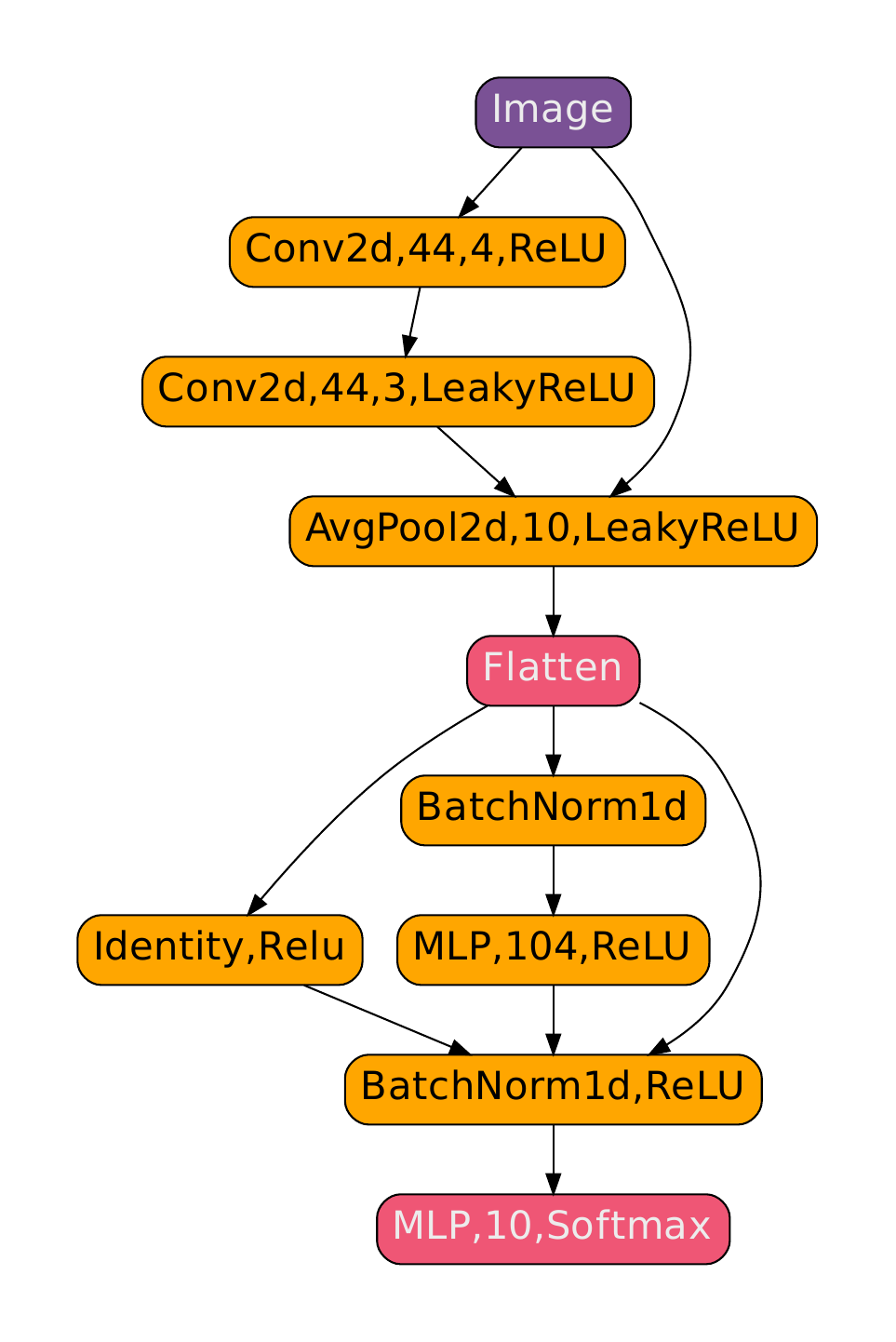}
    \caption{Best model found for CIFAR-10, MRBI and SVHN by Random Search and Hyperband}
    \label{fig:graph_rs_hb}
\end{figure}

Thanks to their evolutionary tools, the evolutionary algorithm and Mutant-UCB where able to create more performing configurations. Figure~\ref{fig:cifar_architectures} shows the models found by both algorithms on the CIFAR-10 data set. The one found by Mutant-UCB, in Figure~\ref{fig:cifar_mutant}, is really close to the original one from the pool displayed Figure~\ref{fig:graph_rs_hb}. The mutation operator was used to add more MLP layers at the end of the neural network which helps improving the model accuracy. On the other hand, the structure shown Figure~\ref{fig:cifar_ssea}, found by the evolutionary algorithm is very different. In this algorithm, a new configuration is created by crossing two parents with the crossover and then applying a mutation to one of the offspring. This double transformation allows to move much further away from the initial pool of solutions. In the case of the CIFAR-10 data set, this led the algorithm to consider very complex structures, which was not necessary to obtain good performance.

\begin{figure}
\centering
\renewcommand\thefigure{(\alph{figure})}
\setcounter{figure}{0}
\begin{subfigure}{}
    \includegraphics[width=0.4\textwidth]{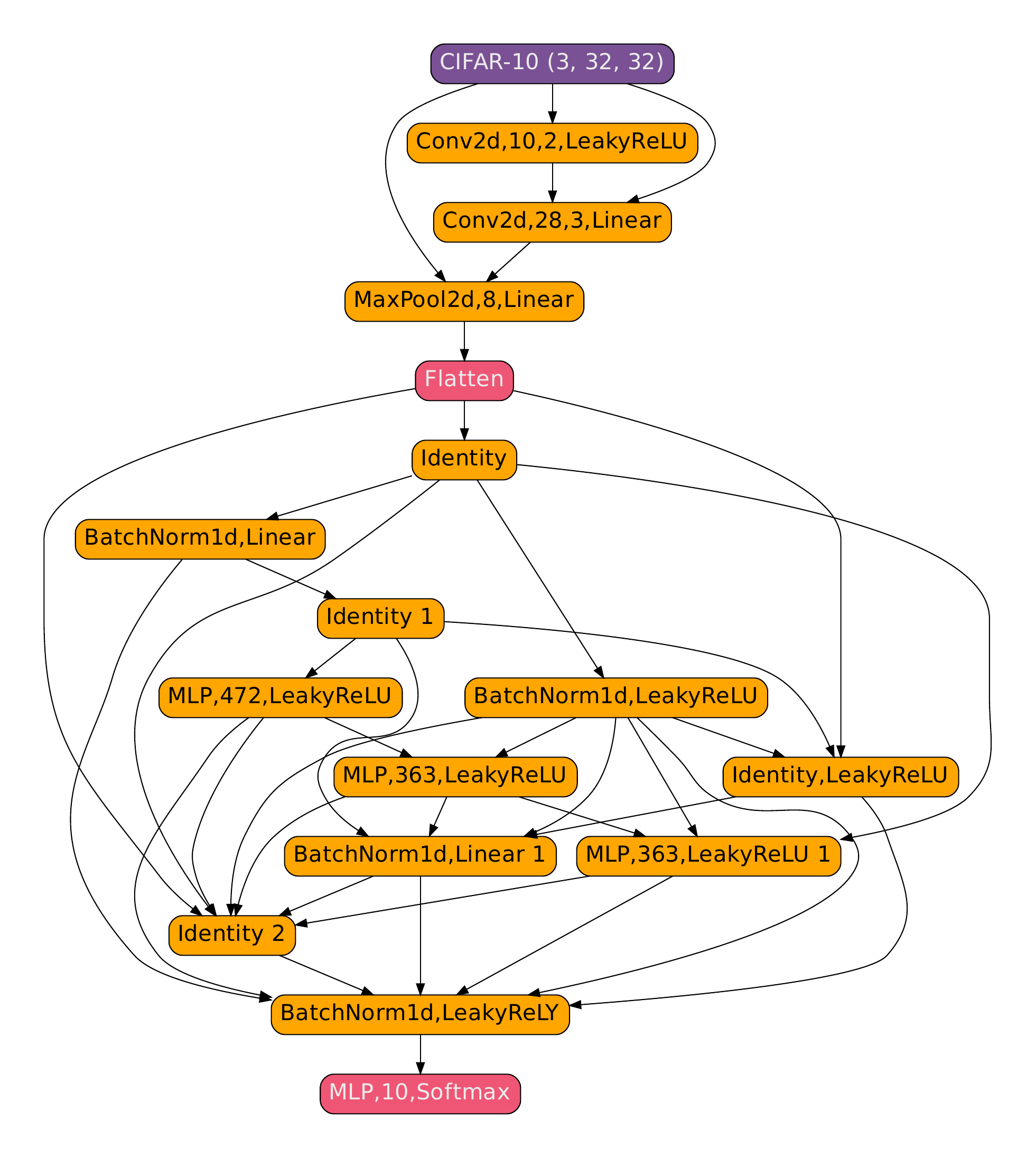}
    \vspace*{-4mm}
    \caption{Evolutionary algorithm.}
    \label{fig:cifar_ssea}
\end{subfigure}
\hfill
\begin{subfigure}{}
    \includegraphics[width=0.4\textwidth]{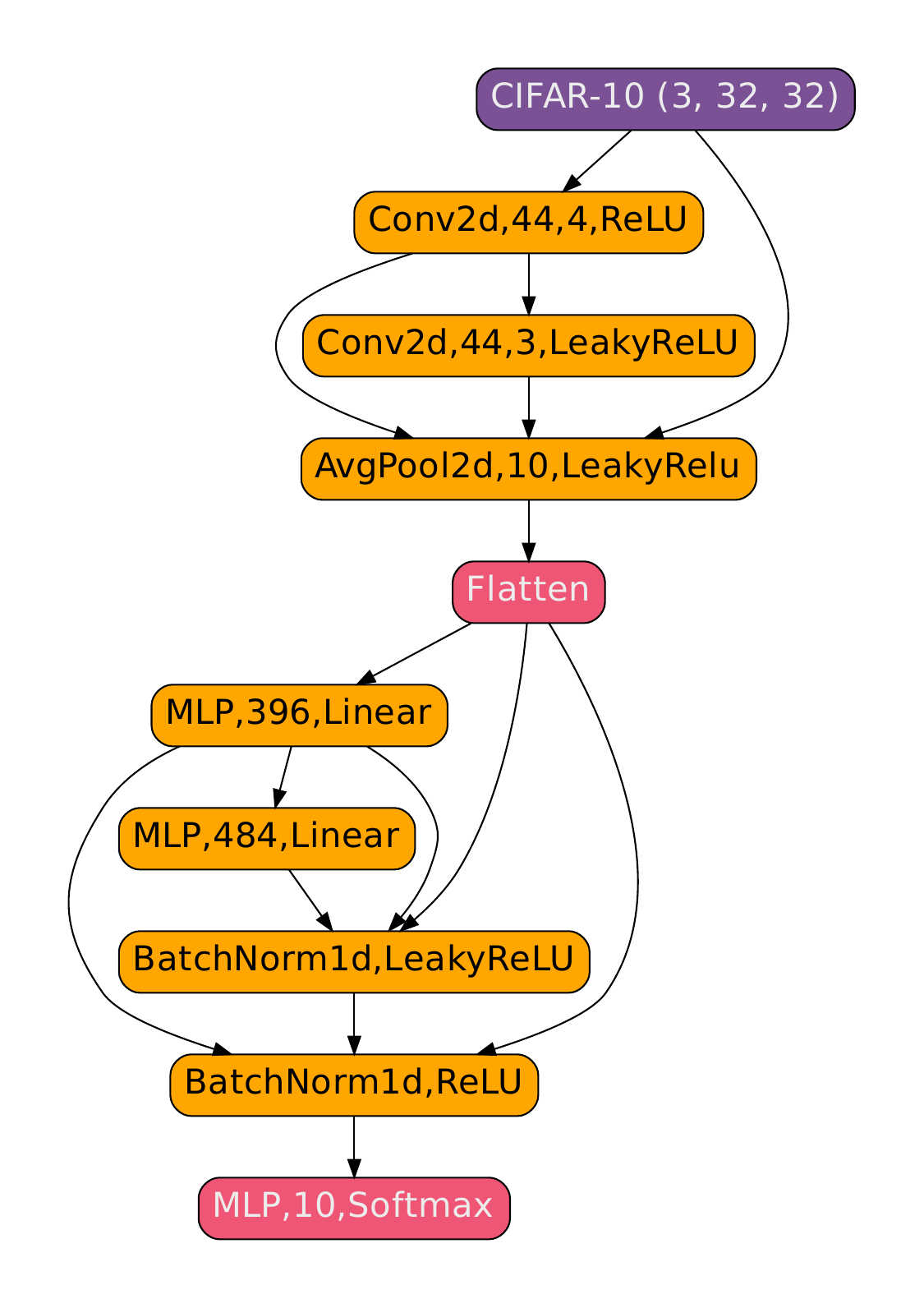}
    \vspace*{-4mm}
    \caption{Mutant-UCB.}
    \label{fig:cifar_mutant}
\end{subfigure}
\setcounter{figure}{6}
\renewcommand\thefigure{\arabic{figure}}
\caption{Best configurations found by the evolutionary algorithm and Mutant-UCB on the CIFAR-10 data set.}
\label{fig:cifar_architectures}
\end{figure}

\end{document}